# Feature identification for parameter extraction and defect detection using machine learning

Y. Guo[1], H. Pahlavani[1], A. Khachaturiants[1], K. Elsayed[1], J. van de Laar[1], E. Simons[1], N. Saikumar[1], H. Sadeghian[1]

[1]Nearfield Instruments, B.V. (The Netherlands)

## ABSTRACT

Process control of advanced semiconductor nodes is not only pushing the limits of metrology equipment requirements in terms of resolution and throughput but also in terms of the richness of data to be extracted to enable engineers to finetune the process steps for increased yield. The move towards 3D structures requires extraction of critical dimension parameters from structures which can vary largely from layer to layer. For in-line process control, the necessary automation forces the development of layer and equipment-specific dedicated image processing algorithms. Similarly, with the increase in stochastic defects in the EUV era, detection of defects at the nm scale requires the identification of features captured in low resolution to meet the throughput requirements of HVM fabs, which can again lead to custom algorithm development. With the emergence of ML-based image processing methods, this process of algorithm development for both cases can be accelerated. In this work, we provide the general framework under which the images obtained from high-speed scanning probe microscopy-based systems can be used to train a network for either feature detection for parameter extraction or defect identification.

## 1. INTRODUCTION

Processing of images obtained from any metrology and inspection equipment in the production line is a crucial step towards extracting relevant information in time for process control. With the reduction of feature sizes and the need for sub-nm resolution 3D imaging, the use of scanning probe microscopy (SPM) is becoming more ubiquitous. This has been accelerated further with the introduction of high-speed SPM metrology and inspection tools like QUADRA which can achieve pixel rates as high as 1MPx/s rivalling other high-speed techniques. Due to prior use of SPM methods mainly in the R&D phase rather than for in-line control, the automatic processing of images obtained using this method is lacking compared to other tools. Additionally, due to the richness of the 3D data obtained with SPM, dedicated image processing methods are necessary for the robust extraction of relevant critical dimensions.

With the high pixel rates now achieved with SPM, and an operator-definable pixel resolution, it additionally offers itself as a viable completely non-destructive defect inspection method for all current and future EUV layers. Especially detection of surface defects such as bridges, missing lines, and missing or connecting holes can be performed in a completely non-destructive fashion using QUADRA, with the tool providing the capability for the review of all found defects through the capture of the full 3D profile at the location of interest. The detection of features, however, again becomes critically important. The problem arises mainly due to the fact that when the measurement parameters are setup for very fast surface scans of high-aspect ratio deep structures, the full depth is not captured and these structures are captured with shallow depths which is sufficient from the perspective of detecting defects. The large possible range of defects additionally creates a heavy burden on the development of algorithms which detect defects with zero false negatives and minimal false positives.

The use of ML-based methods has been extended heavily in semiconductor fabs and these can be used to also reduce the development time of algorithms. In this work, we provide the general framework through which ML models can be developed for the purpose of either feature detection for parameter extraction in the metrology step or for defect identification. A general hindrance to the application of the ML approach is the need for a large dataset, which is common for the training of Neural network (NN) models. To overcome this, we introduce and use a synthetic image generator which has been customized for the creation of SPM images with artifacts from SPM measurements noted in literature placed in images to create a rich and diverse training set. This generator allows for preliminary training of the ML model without the need for real data beyond which we leverage the advantage of transfer learning where the ML model training is refined using real images collected on the SPM tool.

## 2. SYNTHETIC IMAGE GENERATOR

The example of an EUV contact hole layer in the ADI condition is considered to showcase the work in this paper. Apart from full metrology, as noted earlier, defect inspection with the identification of missing or merging contact holes is a critical need with the increase in stochastic defects and shrinking lateral dimensions. In this section, we introduce our customer synthetic image generator, which can create a diverse and rich set of images for training the ML model.

The synthetic image generator consists of two parts as shown in Figure 1. The first one is called the simulator, which is responsible for the creation of noise-free SPM images where differences with structures introduced into the images are related to the layer of interest and the potential stochastic variations & LCDU expected. The simulator provides the user full freedom to determine the following parameters for the generated images:

- Pixel size along both axes of the image
- Feature shape, size, and pitch from the layer
- LCDU resulting in a random distribution and variation in the feature profiles including deviations of pitch
- Number of defects introduced into images at random for defect inspection training scenario

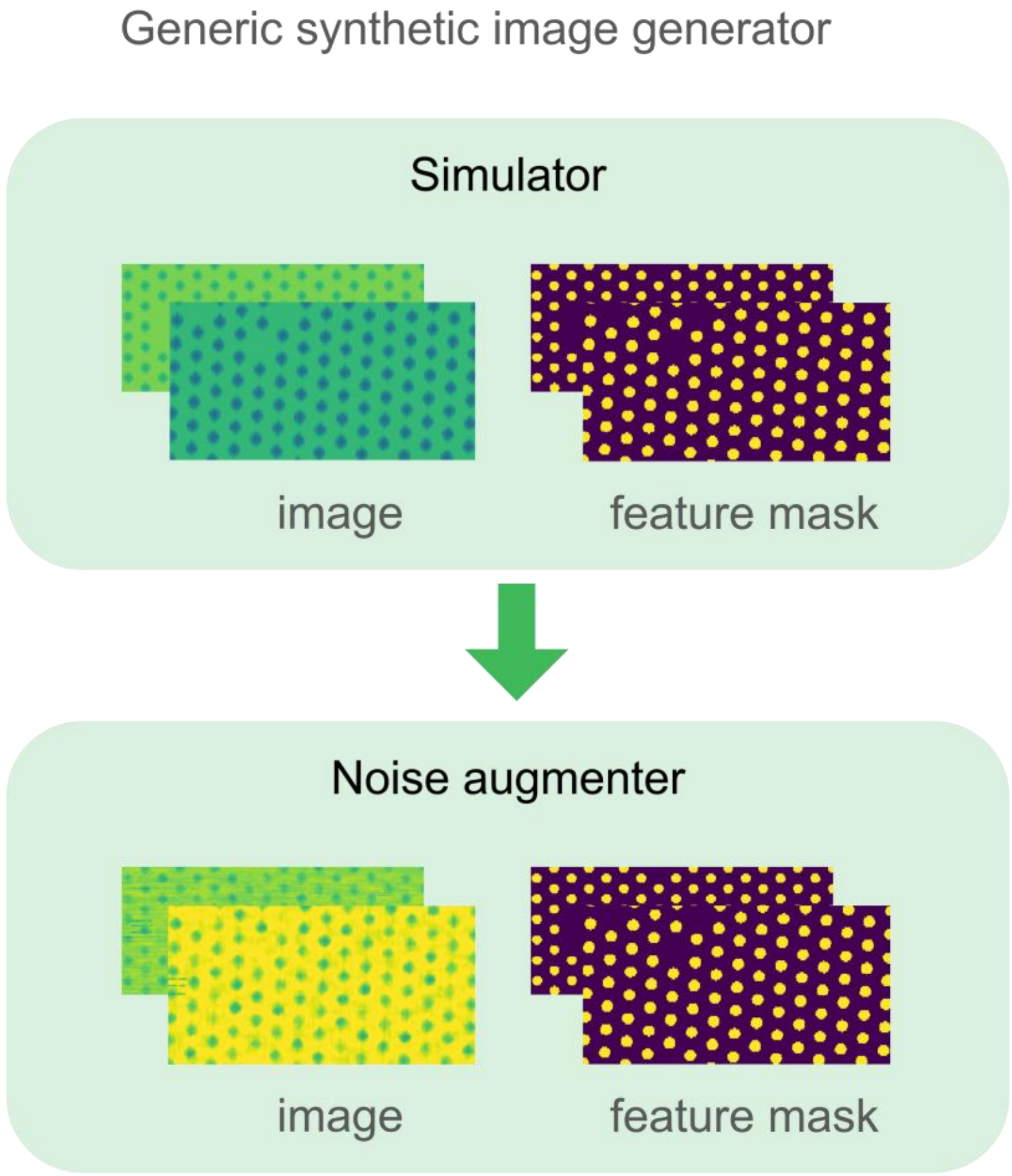

**Figure 1** Synthetic image generator with the simulator creating the initial noise-less image with the corresponding binary segmentation masks and the noise augmenter adding real case scenario noise into the images.

The other part is called the noise augmenter and introduces the most noted noise types from literature into the SPM images including some artifacts coming from the layers. The level of noise introduced into the image can also be controlled by the user specifically aimed at improving the richness of the data and training the ML model for all scenarios. The different artifacts introduced into the images are:

- Streak/scratch type of artifact
- Phase noise
- Large particles where size limits can be determined by the user
- Additive white Gaussian noise

The biggest advantage of the approach of using a synthetic image generator is the complete avoidance of manual labelling of sets for feature detection in both metrology and defect inspection scenarios. Since the semiconductor features are also created by the simulator, the exact location of defects and features is known, and the images are automatically labelled for later use during training. Apart from the synthetic images, the binary segmentation masks are automatically created by the simulator. Some example synthetic images created for missing contact hole defect inspection case are shown in Figure 2.

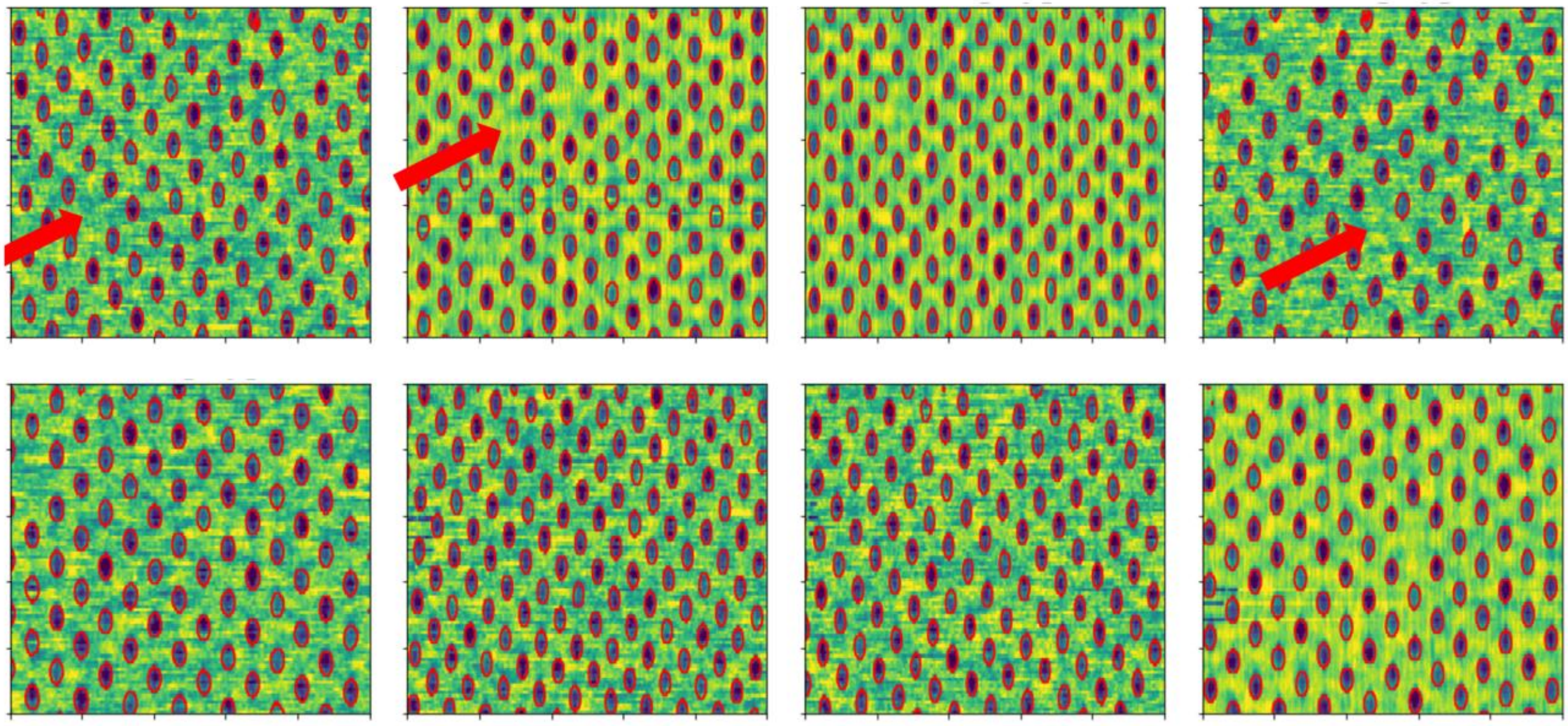

**Figure 2** Example synthetic SPM images of contact hole structures created by the synthetic image generator with random missing contact holes as defects indicated by red arrows. The generator-created segmentation masks are overlayed on the synthetic images.

## 3. UNET FRAMEWORK FOR FEATURE DETECTION

The ML framework used for feature detection is developed using UNet, which has been noted for use in image segmentation problems in literature aimed mainly at medical images [1], though it has recently been extended to SPM images l[2]. The general UNet structure as used in our work is composed of an encoder and a decoder (Figure 3). The encoder consists of several convolution operations followed by rectified linear units (ReLU), which introduces the nonlinearity property into a ML model and solves the vanishing gradients issue; the decoder consists of the same number of convolution operations aimed at up sampling. Apart from the path through convolution operations which creates a bottleneck at the bottom of the shown architecture, direct copy paths connect each level from the left encoder side to the right decoder size ensuring precise localization of the necessary features as is needed for this application. For training, the loss function is determined through equally weight-assigned binary cross entropy (BCE) and dice scores. The dice score loss ensures a balance between false positives and true negatives.

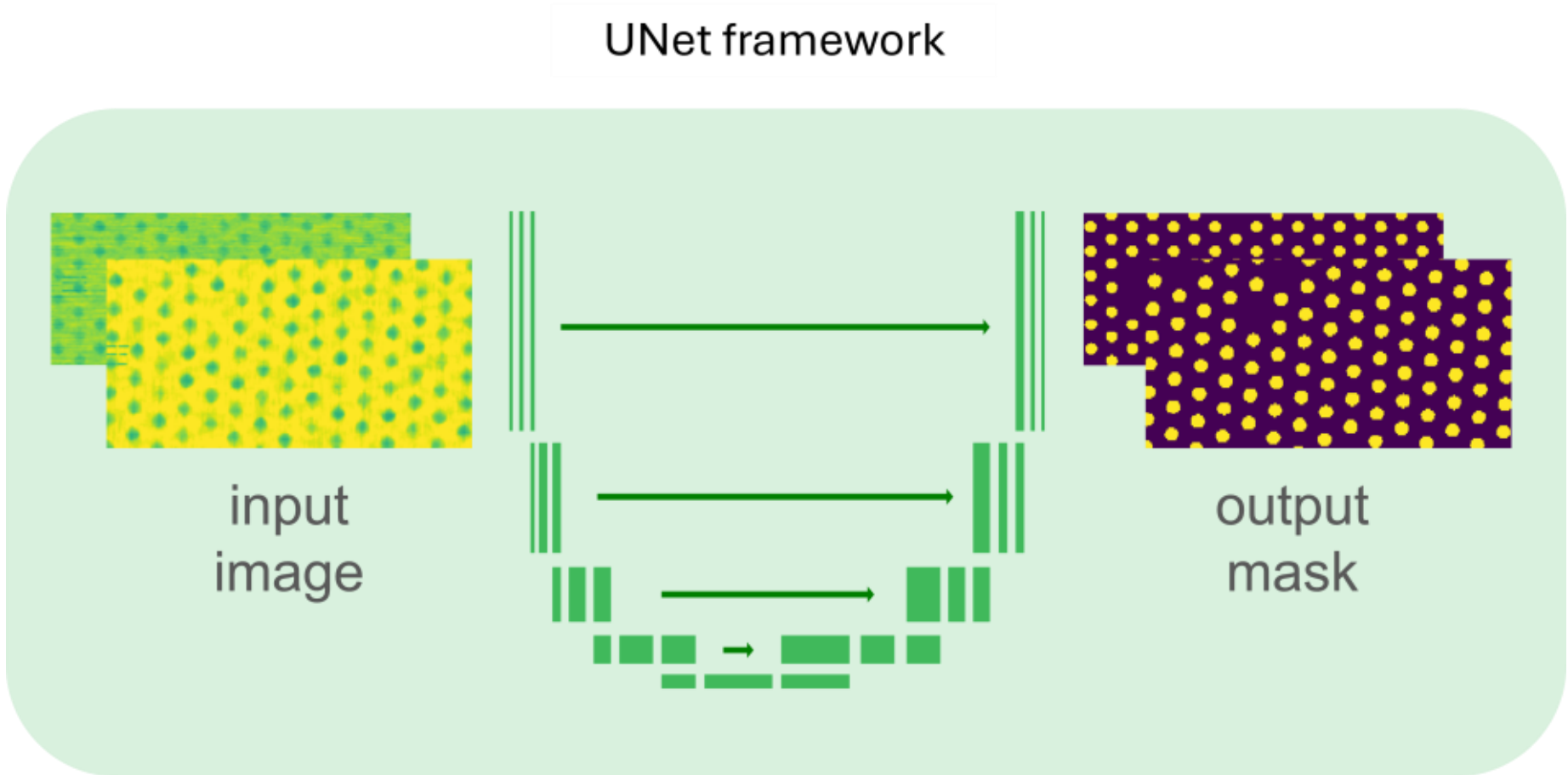

**Figure 3** UNet architecture used for feature detection aimed at both metrology and defect inspection applications.

## 4. RESULTS

The full UNet training parameters used for this example case are provided in Table 1. The trained UNet model is tested on images like those shown in Figure 4 and for the example case of contact hole identification from SPM images, the model provided a100% accuracy in detecting the contact holes with no false positives or negatives. For comparison to traditional image processing methods, a simple thresholding algorithm was developed for the same application and images were processed with both methods. The output of some extreme examples where the contact holes are very shallow leading to improper thresholding application or where scratch-like SPM artifacts are present leading to false positives are shown in Figure 5.

| Parameter | Value |
|---|---|
| Input size | Non-overlapping patches with 64x64 pixels |
| Batch size | 16 |
| Learning rate | 1e-4 |
| Size of convolutional kernel | 3x3 |
| Number of layers | Four layers with bottleneck excluded |
| Number of channels in 1st layer | 64 |
| Activation Function | ReLU |
| Training loss | 0.5 BCE + 0.5 dice loss |
| Optimizer | Adam |
| Epochs | 50 (observed overfitting after 30) |

**Table 1** UNet based model training parameters.

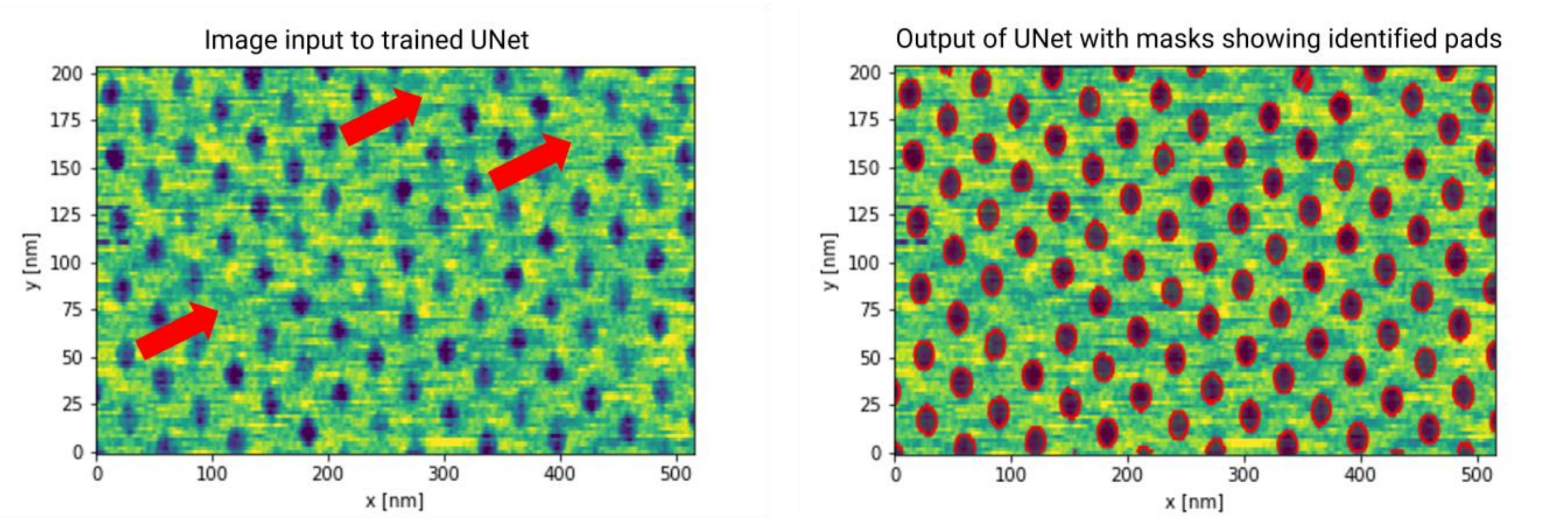

**Figure 4** Example input and output of the trained UNet ML model showing 100% accuracy in identification of contact holes. Missing contact holes (shown through red arrows) can be identified through a comparison of the expected coordinates of holes to those of masks.

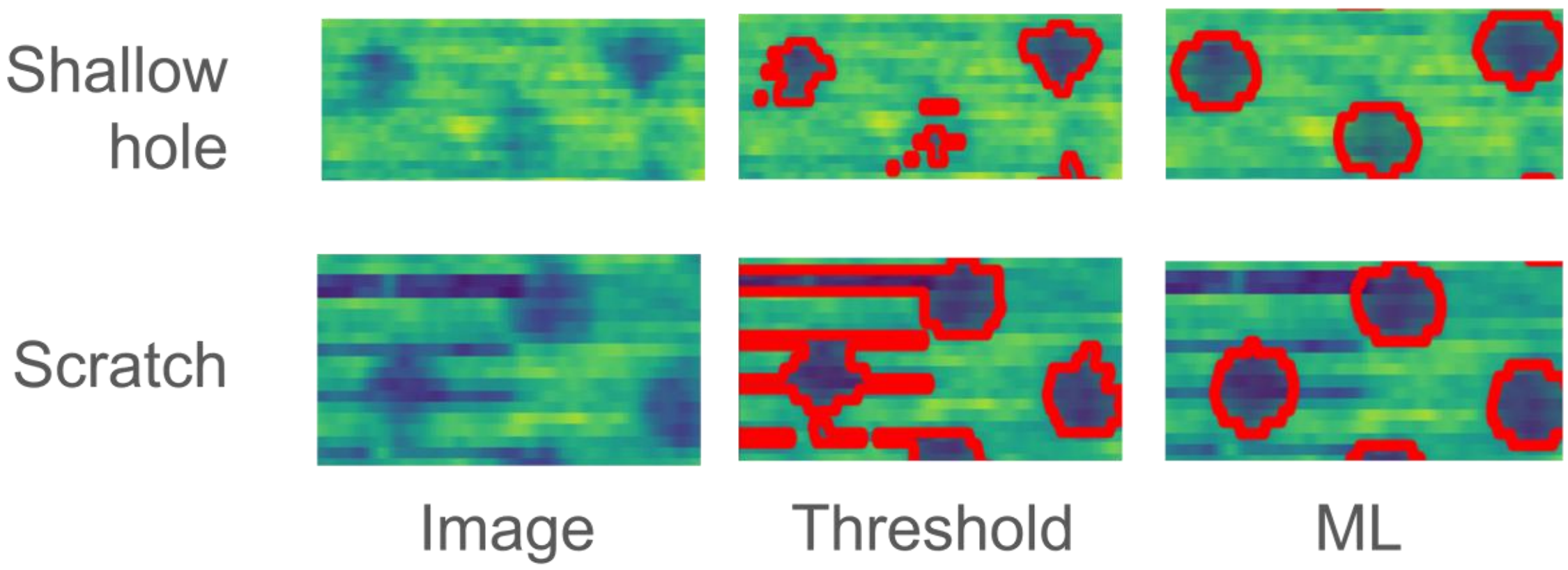

**Figure 5** Comparison of feature detection performance between the developed ML method and a traditional thresholding method used to detect contact holes in cases where the holes are very shallow, or the images are riddled with scratch-like artifacts from SPM.

## 5. CONCLUSIONS

The introduction of high-speed SPM tools for metrology and inspection scenarios in semiconductor fabs necessitates the development of corresponding image processing algorithms aimed at both parameter extraction and defect detection through appropriate feature detection in the captured images. In this work, we present the general framework through which ML-based methods can be developed for this purpose. To overcome the need for a large training dataset, we first introduce a synthetic image generator capable of introducing differences into the noise-less image corresponding to structure variations introduced by the process which is then fed into a noise augmenter converting these noise-less images to contain well-noted SPM related image artifacts and general noise content. These images are then used to train the ML network with the example case of missing contact holes considered in the work. The results from the implementation of the network on the example case are promising and showcase how this framework can be utilized for several SPM related image processing developments.

## ACKNOWLEDGEMENTS

Co-funded by The European Union. The project is supported by the Key Digital Technologies and by RVO (Rijksdienst voor Ondernemend, RVO, The Netherlands). Views and opinions expressed are however those by the authors and do not necessarily reflect those of Key Digital Technologies joint undertaking. Neither the European Union nor the granting authority can be held responsible for them.